\RequirePackage{fix-cm}
\documentclass[twocolumn]{svjour3}          
\smartqed  
\usepackage{graphicx}
\usepackage{cite}
\usepackage{multirow}
\usepackage{amsmath}
\usepackage{amsfonts}
\usepackage{mathrsfs}
\usepackage{graphicx}
\usepackage[font=footnotesize]{subfig}

\usepackage{amsmath}
\usepackage{amssymb}
\usepackage{multirow}
\usepackage{graphicx}
\usepackage{cite}

\usepackage{tikzpagenodes}
\usepackage{lipsum}

\usepackage{algorithm}
\usepackage[noend]{algpseudocode}
\usepackage{url}
\usepackage{textcomp}

\usepackage{hyperref}

\begin{document}

\title{The Achievement of Higher Flexibility in Multiple Choice-based Tests Using Image Classification Techniques}

\titlerunning{Higher Flexibility in MCQ Tests Using Image Classification Techniques}        

\author{ 
Mahmoud Afifi \and Khaled F. Hussain
}

\authorrunning{Mahmoud Afifi and Khaled F. Hussain} 

\institute{ 
 M. Afifi \at 
Electrical Engineering and Computer Science Department, Lassonde School of Engineering, York University, Canada.
\\
Information Technology Department, Faculty of Computers and Information, Assiut University, Egypt
\email{mafifi@eecs.yorku.ca - m.afifi@aun.edu.eg} \and  K. F. Hussain \at Computer Science Department, Faculty of Computers and Information, Assiut University, Egypt. \email{khussain@aun.edu.eg}}

\date{Received: date / Accepted: date}

\maketitle

\begin{abstract}
In spite of the high accuracy of the existing optical mark reading (OMR) systems and devices, a few restrictions remain existent. In this work, we aim to reduce the restrictions of multiple choice questions (MCQ) within tests. We use an image registration technique to extract the answer boxes from answer sheets. Unlike other systems that rely on simple image processing steps to recognize the extracted answer boxes, we address the problem from another perspective by training a machine learning classifier to recognize the class of each answer box (i.e., confirmed, crossed out, or blank answer). This gives us the ability to deal with a variety of shading and mark patterns, and distinguish between chosen (i.e., confirmed) and canceled answers (i.e., crossed out). All existing machine learning techniques require a large number of examples in order to train a model for classification, therefore we present a dataset including six real MCQ assessments with different answer sheet templates. We evaluate two strategies of classification: a straight-forward approach and a two-stage classifier approach. We test two handcrafted feature methods and a convolutional neural network. In the end, we present an easy-to-use graphical user interface of the proposed system. Compared with existing OMR systems, the proposed system has the least constraints and achieves a high accuracy. We believe that the presented work will further direct the development of OMR systems towards reducing the restrictions of the MCQ tests.
\keywords{MCQ \and Optical Mark Reading \and OMR \and CNN \and BoVW \and Dataset of Exams.}
\end{abstract}

\section{Introduction}
\label{intro}
Multiple choice question (MCQ) tests are considered a straight-forward form of written assessment. It is the easiest way to grade, especially for large numbers of students. It can contain a large number of questions that enable covering most of the subject's content. Despite the flexibility of other alternatives of assessment techniques, such as modified essay questions, the MCQ is still deemed the most reliable technique to avoid any misinterpretation of answers \cite{gronlundassessment, mccoubrieimproving}. Thus, many automated MCQ grading systems, also known as optical mark reading (OMR), have been presented. Accordingly, various methods have been proposed for automated generation of MCQs \cite{aldabesemantic, liuautomatic}.

The automated MCQ assessment systems can be summarized in four main steps. First, is obtaining the answer sheet in a digital form, usually done by scanners. Recently, some techniques use digital cameras instead to reduce the cost of the grading systems. Second, is the extraction of student ID and answer boxes. Eliminating this problem, to the best of our knowledge, could be done by using a specific template with the existing systems, called the optical answer sheet or scantron sheet. Third, is the recognition of the student ID which is done using either specific patterns \cite{spadaccinimultiple, chaiautomated} or OCR techniques \cite{fisteusgrading}. Recognizing the student ID is outside the scope of this paper, however many OCR techniques have been presented in the literature, starting from relying on simple image processing techniques \cite{ahmedocr} to more sophisticated techniques, such as convolutional neural network (CNN) solutions \cite{wangend, jaderbergreading}. Fourth, is the recognition of students' responses which is usually performed using simple image processing techniques with the assumption of getting a single dark answer box for each question \textemdash assuming there are no crossed out answers \cite{chinnasarnimage, nguyenefficient, chaiautomated}. This highlights some of the restrictions of the existing MCQ tests. As a result, there are a set of instructions that must be followed by both the student and the examiner in order to use the machine-read answer sheet.

 Fixed templates are a common limitation, as most optical mark recognition devices and systems require a particular form to detect and extract the answer boxes. And so, students tend to struggle with the tiny fonts that are usually used in those templates, in addition to the difficulty of finding the associated answer box for each question. Finally, most of the grading systems consider any filled answer box as a confirmed answer, regardless of whether this answer box is crossed out or not \cite{chaiautomated, chinnasarnimage, hussmannhigh, denglow, levimethod, chouvatutflexible}. This results in students having to ensure that their canceled answers are completely erased to avoid wrong grade reductions.

A few systems adopted a more sophisticated thresholding process giving students the ability to cancel their answers by considering cross signs as confirmed answers and the completely filled boxes as crossed out answers \cite{spadaccinimultiple}. However, they are still restricted with a particular mark pattern.

Our main focus in this work is on increasing the flexibility of dealing with crossed out and confirmed answers with different shades and patterns. To that end, the problem is addressed from another perspective to reduce this constraint of the MCQ tests. We use machine learning (ML) techniques to handle the recognition process. We treat the issue of answer box recognition as a classification problem to successfully distinguish between a crossed out answer and a confirmed answer. Hence, we train a ML classifier to learn the different features among three classes: confirmed, crossed out, and empty (i.e., blank) answers. 

In efforts to get rid of the fixed template issue, we use a simple strategy. First, we assume that all answer boxes, also referred to as regions of interest (ROIs), are given for the model answer sheet only --  this process can be easily determined manually, since it is only for a single sheet, or by employing an object detection model to extract the ROIs. Then, we use an image registration method to detect answer boxes from all student answer sheets. Although the ROI extraction process is not accurate as the predetermined template used by other systems, the trained classification models can recognize the answer boxes with a high accuracy rate. 

\begin{figure*}
\centering
\includegraphics[width=\linewidth]{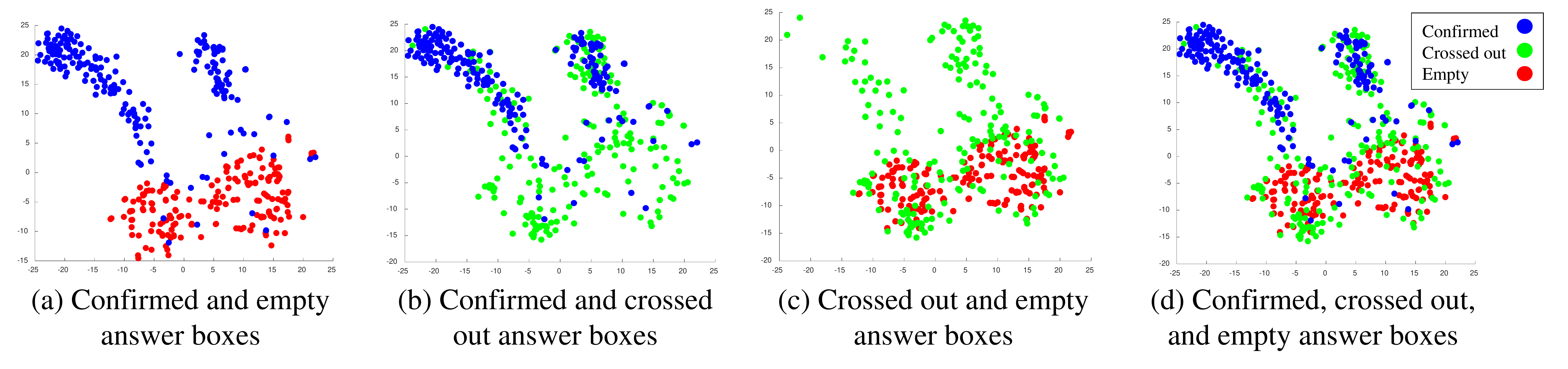}
\caption{The t-distributed stochastic neighbor embedding (t-SNE) visualization \cite{maaten2008visualizing} of 200 random samples of confirmed, crossed out, and empty answer boxes (grayscale) of our proposed dataset. As shown, the distribution of the crossed out answers overlaps the distribution of confirmed and empty answer boxes which leads to low classification accuracy using a simple threshold operation. By considering any answer box as a confirmed answer, if the number of black pixels of the binary version of the given ROI was greater than $T=0.5$, we got 30.83\% accuracy. The binarization process is performed based on the optimal threshold proposed by Otsu \cite{otsuthreshold}.}
\label{fig0}
\end{figure*}

This use of ML classification techniques appears to be largely unexplored in the context of MCQ assessment systems. This is justified, however, as all ML-based classification techniques require a large number of training samples in order to classify the data belonging to different classes. At this point, we collect a dataset of six real MCQ-based assessments with 10,980 confirmed answers, 202 crossed out answers, and 22,367 empty answer boxes. This data was used to train two handcrafted feature classifiers and one deep CNN. Fig. \ref{fig0} shows the insufficiency of using the simple thresholding process to recognize the three classes of answer boxes. Compared with other OMR systems, the proposed system is deemed reliable with a higher flexibility than existing systems. 
\\

\noindent
\textbf{Contribution:}
\\
The main contributions of our work are: (i) a new dataset that could help to revisit the OMR problem again aiming to reduce restrictions of the current OMR systems, and (ii) showing that by treating the problem as a classification problem, we achieve good accuracies without need to a precise extraction of answer boxes' coordinates, as required by other OMR systems. In addition, we show that by benefiting from the ML techniques, we can handle the case of crossed out answers.

The rest of the paper is organized as follows: Section \ref{related} briefly reviews some related research methods and systems. In Section \ref{existing}, we study four of the present mobile applications for the MCQ test assessment. Section \ref{method} outlines the methodology. Then, we present the proposed dataset of real MCQ tests in Section \ref{dataset}, followed by the experimental results in Section \ref{results}. The paper is concluded in Section \ref{conclusion}.

\noindent

\section{Related Work}
\label{related}

The OMR is the main block for many applications, such as the paper-based ballot systems \cite{new1, new2, new3, new4, new5}. In this section, we briefly review the existing MCQ grading systems that are related to our work.

Many traditional OMR devices, with linked scanner devices, are used to grade MCQ tests. Predetermined locations of answer boxes are scanned and based on the reflectivity of these answer boxes, the automatic grading is performed \cite{smithoptical, hussmannhigh, loprestidocument}. Regular image scanners are used in OMR software applications \cite{sanguansatrobust}. Recently, low-cost systems relying on web-cams \cite{fisteusgrading} or digital mobile cameras \cite{chinaapplication} were presented. Interestingly, when the used approaches for answer box recognition were reviewed, it was discovered that the recognition process was typically performed using simple image processing operations.

The early OMR system \cite{chinnasarnimage} relied on the histogram of binary image pixels to recognize the answer boxes. Hussmann and Deng \cite{hussmannhigh} developed a high-speed OMR hardware which can deal with three different mark shapes (circle, oval, and rectangle). Since the thresholding technique has a low computational cost, a simple threshold operation in a single-chip field programmable gate
array had been implemented to recognize the predetermined answer boxes.  
The same simple strategy was followed by Deng et al. \cite{denglow} who used a thresholding process based on the black pixels in the binary image after a set of pre-processing steps. 
Nguyen et al. \cite{nguyenefficient} presented a low-cost system based on a digital camera instead of scanners in the image acquisition process. Accordingly, they have applied skew adjustment and normalization steps to locate the answer boxes. Eventually, the decision is made based on the number of black pixels.
Spadaccini et al. \cite{spadaccinimultiple} presented a new term, called squareness, to the thresholding operation, which is the absolute difference between the number of black pixels in the columns and rows. This helps deal with crossed out answer boxes by recognizing the cross signs as confirmed answers and completely black boxes as canceled answers.

J. A. Fisteus et al. \cite{fisteusgrading} proposed a system called \textit{Eyegrade} which is based on web-cams. The \textit{Eyegrade} entails a two-stage thresholding method to recognize the answer boxes. This allows the system to handle the canceled answers by following the same strategy proposed in \cite{spadaccinimultiple}; where the student has to completely fill the canceled answer boxes.

Sanguansat \cite{sanguansatrobust} proposed a more complex recognition technique, using adaptive thresholding to improve the robustness through dealing with different mark patterns. However, the accuracy is insufficient (85.72\%) for a specific pattern. D. Chai \cite{chaiautomated} presented a semi-template-free MCQ test assessment system. The system uses a fixed sheet structure that has specific areas for answer boxes, instructions, and student IDs to facilitate the process of detecting the important data while grading. Eventually, the decision is made based on the number of black pixels in the answer box. Also, there are a set of mobile applications that are available to grade MCQ tests automatically.

\begin{figure*}
\centering
\includegraphics[width=0.75\linewidth]{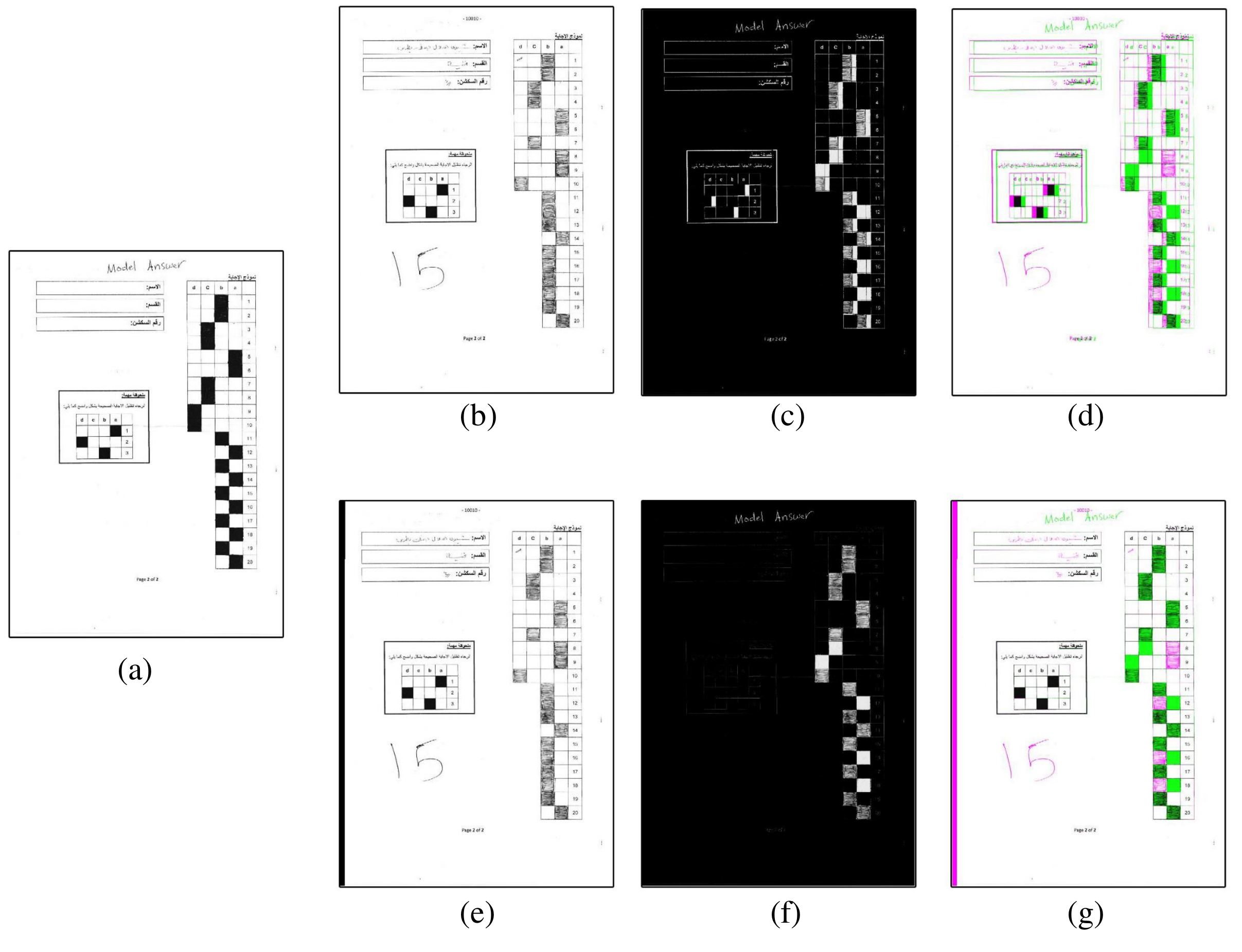}
\caption{The alignment process of answer sheet image (b) with reference image (a). The difference between both images is shown in (c) and the overlap visualization is shown in (d). After applying the registration process, the registered image (e) is aligned with the reference image; that leads to small differences between images as shown in (f) and (g).}
\label{fig2}
\end{figure*}

\begin{figure*}
\centering
\includegraphics[width=\linewidth]{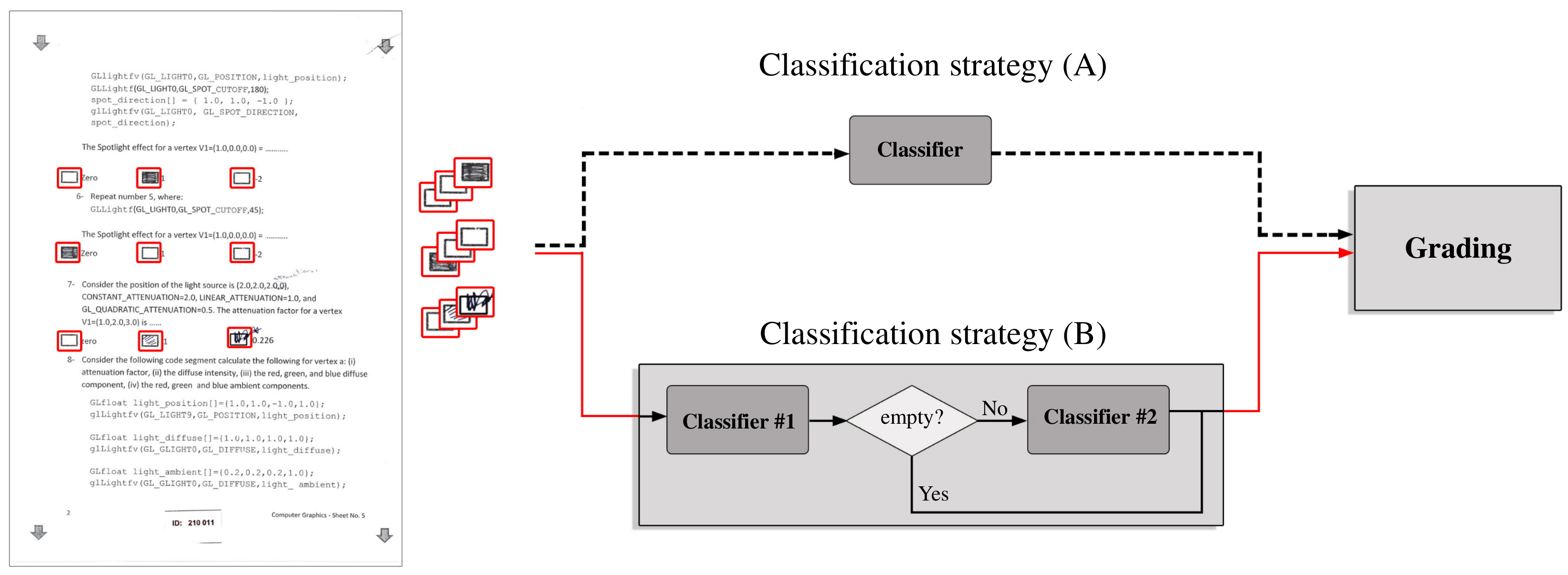}
\caption{Overview of the examined classification strategies. Classification strategy (A) is the straight-forward way to classify the three different classes. The second strategy (B) divides the process into two stages. The first one is performed by classifier number 1 to distinguish between confirmed and blank answers. If the answer box is classified as confirmed, the second classifier double checks the answer box as either a confirmed or a crossed out answer.}
\label{fig4}
\end{figure*}

\section{Accuracy of Existing Mobile Applications}
\label{existing}
In order to understand the main drawbacks of the existing mobile applications, we have examined four MCQ grading mobile applications \footnote{(1) MCTest Corrector. (2) ZipGrade. (3) MCScanner. (4) Exam Reader (ER).}. We have applied two different tests for each mobile application. The first test was the simplest one; there were 20 different answer sheets graded with adherence according to the marking instructions given by each application. The second one was more complicated using crossed out answers, and different shading and pattern styles of confirmed answers. Each one of them gives a different accuracy under different lighting conditions and camera resolutions. 

The average accuracy obtained in the first test is 87.7\%; while, the average accuracy of the last test was 70.2\%. Moreover, all of the mobile applications mentioned above need a fixed template to be able to grade and do not deal with crossed out answers. As far as we know, the technique used by these applications has not been published. Hence, we could not mention the technical details of the recognition process.  

\begin{table*}
\centering
\caption{Summary of the MCQ tests in the proposed dataset}
\label{table1}
\begin{tabular}{l|c|c|c|c|c|c|c|}
\cline{2-8}
\multirow{2}{*}{} & \multicolumn{7}{c|}{ExamID} \\ \cline{2-8} 
 & Exam0 & Exam1 & Exam2 & Exam3 & Exam4 & Exam5 & Total \\ \hline
\multicolumn{1}{|l|}{\begin{tabular}[c]{@{}l@{}}Number of answer sheets\end{tabular}} & 40 & 99 & 103 & 60 & 62 & 371 & \textbf{735} \\ \hline
\multicolumn{1}{|l|}{\begin{tabular}[c]{@{}l@{}}Pages per answer sheet\end{tabular}} & 3 & 1 & 1 & 1 & 1 & 1 &  \\ \hline
\multicolumn{1}{|l|}{\begin{tabular}[c]{@{}l@{}}Number of confirmed answers\end{tabular}} & 519 & 979 & 1,041 & 599 & 551 & 7,291 & \textbf{10,980} \\ \hline
\multicolumn{1}{|l|}{\begin{tabular}[c]{@{}l@{}}Number of crossed out answers\end{tabular}} & 16 & 24 & 17 & 11 & 42 & 92 & \textbf{202} \\ \hline
\multicolumn{1}{|l|}{\begin{tabular}[c]{@{}l@{}}Number of empty answers\end{tabular}} & 1,034 & 1,967 & 2,032 & 1,190 & 1,267 & 14,877 & \textbf{22,367} \\ \hline
\multicolumn{1}{|l|}{\begin{tabular}[c]{@{}l@{}}Total number of pages\end{tabular}} & 120 & 99 & 103 & 60 & 62 & 371 & \textbf{815} \\ \hline
\end{tabular}
\end{table*}

\section{Methodology}
\label{method}
The proposed system assumes that the ROIs for each question in the model answer have been given. For each answer sheet, the proposed system extracts the ROIs in a semi-automated manner. Then, the class of each ROI is classified as a confirmed, a crossed out, or an empty answer box. For each question, we seek to find the confirmed answer. If two confirmed answers are found, the mark will be zero for this question. Crossed out answers are ignored, except in the case that they are found without other confirmed answers in a given question, see Algorithm \ref{algo1}.

\begin{algorithm}
\caption{The proposed grading algorithm}\label{algo1}
\begin{algorithmic}[1]
\Procedure{grade}{answer\_sheet, metadata, classification\_strategy}
\State $\textit{I} \gets \text{answer\_sheet}$
\State $\textit{M} \gets \text{metadata}$
\State $\textit{G} \gets 0$
\State $\textit{s} \gets \text{classification\_strategy}$ \Comment{$\text{Strategy: 1 or 2}$}
\State $j \gets 1$
\While {$j<=numberOfQuestions(metadata)$}
\State $answers \gets 0$
\State $ROIs \gets \textit{extractROI}(I,M,j)$
\State $i \gets 1$
\While {$i <= \textit{length(ROIs)}$} 
\State $c(i) \gets \textit{classify}(ROIs(i),s)$.
\State $check1 \gets \textit{c}(i) == \textit{2}$
\State $check2 \gets answers==0$
\State $check3 \gets \textit{c}(i) == \textit{1}$
\If {$(check1 \text{ AND } check2) \text{ OR } check3$} 
\State $answer \gets i$
\If {$check3$}
\State $answers \gets answers+1$
\EndIf
\EndIf
\State $i \gets i+1$
\EndWhile

\If {$answers<=1$} 
\If {$correct(answer,M,j) == true$} 
\State $G=G+getGrade(M,j)$
\EndIf
\EndIf
\State $j \gets j+1$
\EndWhile
\Return $G$
\EndProcedure
\end{algorithmic}
\end{algorithm}

\subsection{Extracting ROIs}
The answer sheets are aligned with the model answer document to avoid any misalignments in the scanning process. Firstly, the answer sheet image is converted to a monochrome image. The speeded up robust features (SURF) \cite{SURF} are extracted and matched with the pre-extracted SURF of the reference image, namely model answer image, using the fast matching technique presented by Muja et al. \cite{matching}. The M-estimator sample consensus (MSAC) algorithm \cite{MSAC} is used to estimate the invertible geometric transform matrix \textbf{\textit{T}} from the matched pair of SURF. The ROIs' corner points of the given answer sheet are transformed to the model answer sheet using the transformation matrix given by
\begin{equation}
{\mathbf{T}}=\bigl(\begin{smallmatrix}
{\mathbf{R}} &\mathbf{d^{T}}\\ 
\mathbf{z} & 1
\end{smallmatrix}\bigr)
\end{equation}
where $\mathbf{R}$ is the ($3\times 3$) rotation matrix, $\mathbf{d}$ is the $(1\times 3)$ translation vector, and $\mathbf{z}$ is a $(1\times 3)$ zero vector. 
\\
Linear interpolation is used instead of the cubic interpolation to speed up the process of applying ${\mathbf{T}}$ to the answer sheet image. Fig. \ref{fig2} shows how the alignment operation fixes the accidental shifting during the scanning process. 
\subsection{Classification Strategies}

We proposed two different classification strategies in order to distinguish among confirmed, crossed out, and empty (blank) answer boxes. As shown in Algorithm \ref{algo2}, the first strategy is a straight-forward one. Where the output of the pre-trained classifier is one of the three aforementioned classes (i.e., three-classes classifier). In the second strategy, we aim to reduce the conflict between the crossed out and the confirmed classes. In other words, the classification problem becomes easier, if we classify between two classes, namely confirmed and empty. That being, we can reduce the classification error by splitting the problem into two stages. The first stage is to classify between filled and empty answer boxes. At this stage, the crossed out answer boxes are to be classified as either confirmed or empty answer. Consequently, we apply a second stage of classification to distinguish between confirmed and crossed out answers. See Fig. \ref{fig4}.

\begin{figure*}
\centering
\includegraphics[width=\linewidth]{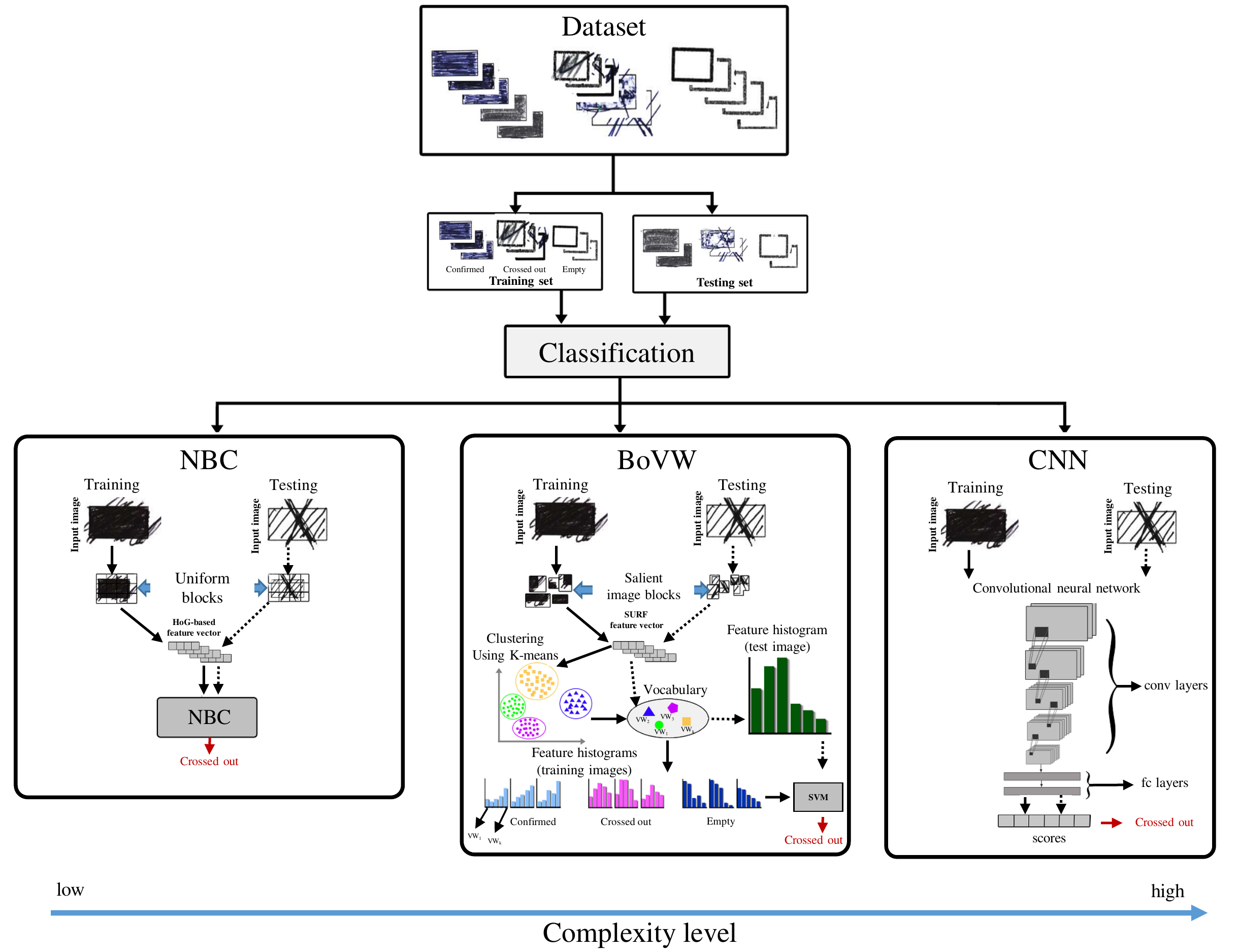}
\caption{Summary of the image classification models used in the answer recognition process. The complexity level increases from left to right in the shown figure.}
\label{summary}
\end{figure*}

\subsection{Classification Methods}
In the literature, many generic classifiers have been presented. The issue with the image classification is how to extract robust features that can feed a generic classifier in order to classify images based on their content. We test three image classifiers. The first one is based on extracting handcrafted features and feeding them to a simple classifier. For simplicity, we adopt the naive Bayes classifier (NBC) \cite{naive} which is a simple generic classifier, used to check whether simple solutions are sufficient to our problem. Next, we adopt a more sophisticated approach for image feature extraction and representation, which is the bag of visual words (BoVW) model \cite{bovw}. The third model is one of the popular deep neural network architectures proposed by Krizhevsky et al.\cite{alexnet}. Fig. \ref{summary} summarizes the three image classification models examined in this work.

\begin{algorithm}
\caption{The proposed classification strategies}\label{algo2}
\begin{algorithmic}[1]
\Procedure{classify}{ROI, classification\_strategy}
\State $\textit{I} \gets \text{ROI}$
\State $\textit{s} \gets \text{classification\_strategy}$
\State $\textit{model} \gets \text{load classification model(s)}$
\State $modelNumber=1$
\State $class \gets \textit{evaluate}(I,model,s,modelNum)$
\If {$\text{NOT } (class==3) \text{ AND } s==2$}  
\State $modelNumber=2$
\State $class \gets \textit{evaluate}(I,model,s,modelNum)$
\EndIf
\Return $class$

\EndProcedure
\end{algorithmic}
\end{algorithm}

\subsubsection{Naive Bayes Classifier}
The NBC is based on a naive but effective assumption, which involves that each feature in the feature vector is independent, i.e., uncorrelated with other features that represent the same input. To customize this generic classifier to our problem, a feature vector was extracted to represent each class, namely confirmed answer, crossed out answer, and empty answer boxes. We use the gradient of the ROI, that is given by
\begin{equation}
g=|\frac{\partial f}{\partial x}| + |\frac{\partial f}{\partial y}|,
\end{equation}  
where $f$ is the given ROI image. The feature vector $\mathbf{v} \in \mathbb{R}^{D}$ is ($1 \times 12$) vector, i.e., $D=12$, which contains the maximum, median, and mean of $g$. The remaining values are based on the histogram of oriented gradients (HoG) \cite{HOG}. In order to get a fixed length HoG feature vector, each given ROI image is resized to a fixed size (227$\times$ 227 pixels). Then, the image is divided into uniform blocks, each one is encoded using 8 bins to eventually get a (1$\times$23328) HoG feature vector.

\begin{figure*}
 \centering
 \includegraphics[width=0.8\linewidth]{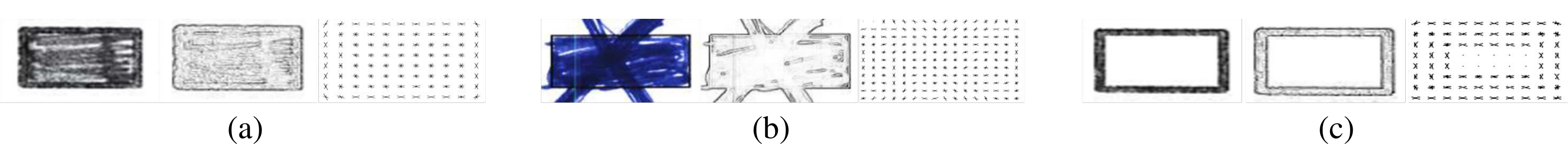}
 \caption{The gradient and histogram of oriented gradients of confirmed answer (a), crossed out answer (b), and empty answer box (c). In (a-c), the true color images are the original ROIs. The second images are the gradient of the corresponding images. The third images are the histogram of oriented gradients.}
 \label{fig5}
 \end{figure*}

To reduce the complexity and the computational time, we extract 9 features from the HoG feature vector which are the mean of: the gradient of the HoG feature vector, the gradient of first bin in all blocks, the gradient of second bin in all blocks, ... etc. The intuitive meaning of the feature vector $\mathbf{v}$ is that the empty box is expected to have no changes except the changes in the border of the answer box. On the other hand, the crossed out and confirmed answer boxes are expected to have a higher amount of changes in the intensity. The HoG feature vector presents the orientations of gradient in the ROI. This may help distinguish between crossed out and confirmed answer boxes; where, the crossed out answers are expected to have a higher orientation of gradient. Fig. \ref{fig5} exhibits the idea behind the feature vector.

The probability of each class given the feature vector of the input image is calculated by
\begin{equation}
P(C|\mathbf{v})=\frac{P(\mathbf{v}|C)p(C)}{P(\mathbf{v})},
\end{equation}
where $P(C|\mathbf{v})$ is the probability of class $C$ given the input vector $\mathbf{v}$. The probability of class $C$ given the input vector $\mathbf{v}$ can be represented as

\begin{equation}
P(C|\mathbf{v})=\frac{\prod_{i=1}^{D} P(F_i|C)p(C)}{\sum_{c'}\prod_{i=1}^{D} P(F_i|C=c')p(C=c')}.
\end{equation}
As shown, we assume that $\mathbf{v}$, in line with the naive Bayes assumption, consists of a set of independent features $F_i, i\in \{1,...,D\}$. The likelihood $P(\mathbf{v}|C)$ is given by the Gaussian distribution; where each feature is represented by a Gaussian distribution with its own mean and variance. The maximum likelihood estimation is used to calculate the mean and variance for each feature. The prior $P(C)$ is represented by a multinoulli distribution. As shown in Table \ref{table1}, the number of crossed out answers is very small compared to the other classes. Hence, we change it to be $5\%$ instead of the real categorical distribution $0.6\%$.

\subsubsection{Bag of Visual Words}
Although the BoVW model may use the NBC or support vector machine (SVM) \cite{svm} at the last stage of the image classification process, it is based on more complex feature representation method. Instead of extracting a simple feature vector, the BoVW model is based on extracting affine invariant features, i.e., features that are robust against affine transformations (rotation, translation, scaling and shearing). The Harris-Affine detector \cite{harr} is used to extract affine invariant features. These features are described using a feature descriptor -- in our case, we have used the SURF descriptor. However, comparing each feature descriptor with all training descriptors leads to a high computational cost. Thus, clustering is performed using the K-means algorithm \cite{duda}. Each cluster center represents a visual word. As a consequence, a pool of visual words (also called vocabulary) is created in the training stage. Then, each image is presented as a bag of visual words, i.e., histogram of visual words in the given image. Finally, a classifier is used to learn the decision boundaries among classes.

In our experiments, we have used a multi-SVM classifier. In the testing stage, the same process is repeated, but the extracted visual features are approximated to the existing visual words that have been generated in the training stage. Fig. \ref{summary} shows the training and testing steps of the BoVW model. As shown, the SURF features are extracted from the answer boxes. These feature vectors are clustered to generate the vocabulary of visual words. For each input answer box, in both training and testing stages, a histogram of visual words is generated to feed the SVM classifier. It should note that the vocabulary is generated only using the training data.

\subsubsection{CNN}
Ordinary neural networks consist of multiple layers of a set of neurons, each of which applies a linear-function followed by a non-linear function to the input vector $x \in \mathbb{R}^D$, where $D$ is the dimensionality of the input vector $x$. The function is given by
\begin{equation}
(f\circ g)(x).
\end{equation}
There are two main components shown in the previous equation which are: (1) the non-linear activation function $f(.)$ and (2) the linear-function $g(.)=w^Tx+b$. The latest applies a simple linear-transformation to the given $x$ using the weight vector $w \in \mathbb{R}^D$ and a bias term $b$. 
The learnable parameters (i.e., $w,b$) are estimated after the training process, using gradient-based optimization techniques \cite{deep}. The CNN is designed to deal with images; certain image-based properties (e.g., convolution operations) are involved in the process to get more precise results. In 2012, the computer vision community paid more attention to CNNs after the noticeable improvement obtained by the AlexNet architecture proposed by Krizhevsky et al. \cite{alexnet}. 
Since then, different CNN architectures have been presented (e.g., VGG \cite{vgg}, ResNet \cite{resnet}) that obtain more accuracy using either a deeper network \cite{vgg} or a more sophisticated forward function \cite{resnet}. In our problem, however, we believe that the AlexNet is sufficient in terms of accuracy and performance \textemdash AlexNet is considered as one of the simplest CNN models in terms of the number of operations and the inference time \cite{canzianianalysis}. Table \ref{AlexNetTable} shows the details of the network layers.

Instead of starting to train from scratch, we have fine-tuned a pre-trained AlexNet model using learning rate $\lambda=0.0001$ for the pre-trained layers and $\lambda=0.002$ for the last fully connected layer. We have used stochastic gradient descent with momentum \cite{murphymachine} to update the learn-able parameter in the training stage for 40 epochs.

\begin{table}
\centering
\caption{The architecture of the CNN model used in this work.}
\label{AlexNetTable}
\scalebox{0.9}{
\begin{tabular}{|l|l|}
\hline
Layer & Description \\ \hline
\textit{input} & \begin{tabular}{p{6.5cm}} $227\times 227\times 3$ images with zero-center normalization \end{tabular}\\ \hline
\textit{conv1 }& \begin{tabular}{p{6.5cm}} 96 $11\times 11\times 3$ convolutions with stride = 4 and no padding \end{tabular}\\ \hline
\textit{relu1} & ReLU activation layer \\ \hline
\textit{norm1} & \begin{tabular}{p{6.5cm}} Cross channel normalization with 5 channels per element \end{tabular}\\ \hline
\textit{pool1} & \begin{tabular}{p{6.5cm}} $3\times 3$ max pooling with stride = 2 and no padding \end{tabular}\\ \hline
\textit{conv2} & \begin{tabular}{p{6.5cm}} 256 $5\times 5\times 48$ convolutions with stride = 1 and padding = 2 \end{tabular}\\ \hline
\textit{relu2} & \begin{tabular}{p{6.5cm}} ReLU activation layer \end{tabular}\\ \hline
\textit{norm2} & \begin{tabular}{p{6.5cm}} Cross channel normalization with 5 channels per element \end{tabular}\\ \hline
\textit{pool2} & \begin{tabular}{p{6.5cm}} $3\times 3$ max pooling with stride = 2 and no padding \end{tabular}\\ \hline
\textit{conv3} & \begin{tabular}{p{6.5cm}} 384 $3\times 3\times 256$ convolutions with stride = 1 and padding = 1 \end{tabular}\\ \hline
\textit{relu3} & \begin{tabular}{p{6.5cm}} ReLU activation layer \end{tabular}\\ \hline
\textit{conv4} & \begin{tabular}{p{6.5cm}} 384 $3\times 3\times 192$ convolutions with stride = 1 and padding = 1 \end{tabular}\\ \hline
\textit{relu4} & \begin{tabular}{p{6.5cm}} ReLU activation layer \end{tabular}\\ \hline
conv5 & \begin{tabular}{p{6.5cm}} 256 $3\times 3\times 192$ convolutions with stride = 1 and padding = 1 \end{tabular}\\ \hline
\textit{relu5} & \begin{tabular}{p{6.5cm}} ReLU activation layer \end{tabular}\\ \hline
\textit{pool5} & \begin{tabular}{p{6.5cm}} $3\times 3$ max pooling with stride = 2  and no padding \end{tabular} \\ \hline
\textit{fc6} & 4096 fully connected layer \\ \hline
\textit{relu6} & ReLU activation layer \\ \hline
\textit{drop6} & 50\% dropout layer \\ \hline
\textit{fc7} & 4096 fully connected layer \\ \hline
\textit{relu7} & ReLU activation layer \\ \hline
\textit{drop7} & 50\% dropout layer \\ \hline
\textit{fc }& 3 fully connected layer \\ \hline
\textit{softmax} & Softmax layer \\ \hline
\end{tabular}
}
\end{table}

\subsection{Graphical User Interface}
We designed an easy-to-use graphical user interface (GUI) of the proposed system as shown in Fig. \ref{fig6}. First, the user identifies the model answer sheet followed by manual determination of the metadata (e.g., all possible answer boxes). The GUI gives the user the ability to choose the classification strategy and the classifier. Finally, the user has to determine the directory where the images of the answer sheets are located in. The system automatically grades all answer sheets and generates a report that contains the grade of each corresponding image file. The report can be rendered in three different formats: (1) XML format (.XML), (2) comma-separated values (.CSV), and (3) Microsoft Excel workbook (.XLSX). In CSV, we represent the image names as a sequence of numbers. Each answer sheet is represented by the image file name. The source code and the trained models are available in the following link: \href{https://goo.gl/XxLWRi}{https://goo.gl/XxLWRi}.

\section{Dataset}
\label{dataset}

\begin{figure*} 
\centering
\includegraphics[width=\linewidth]{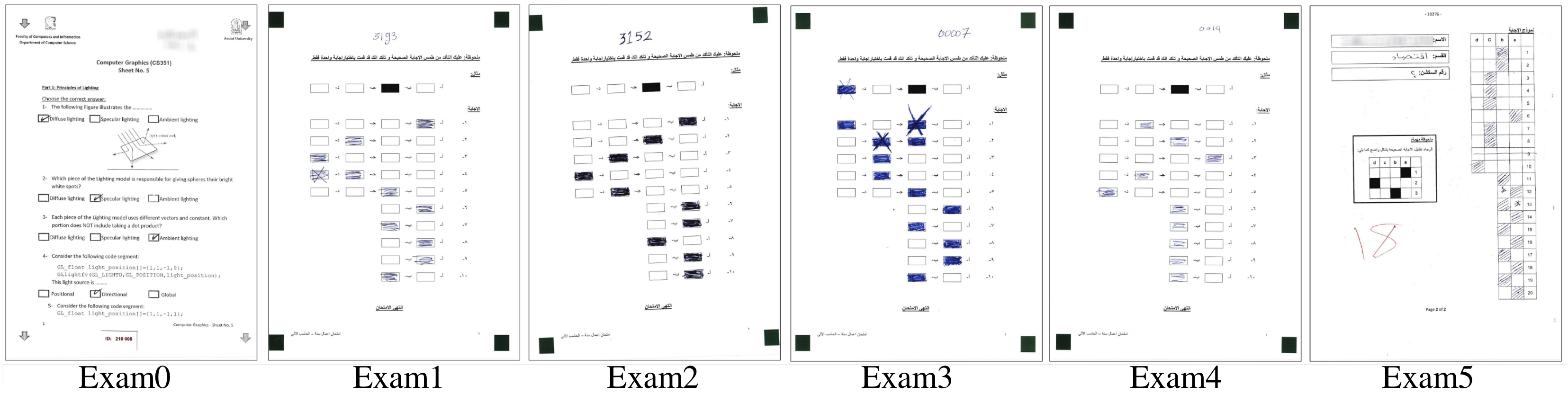}
\caption{Samples of answer sheets in the proposed dataset.}
\label{fig1}
\end{figure*}
\noindent

\begin{figure*}
\centering
\includegraphics[width=0.7\linewidth]{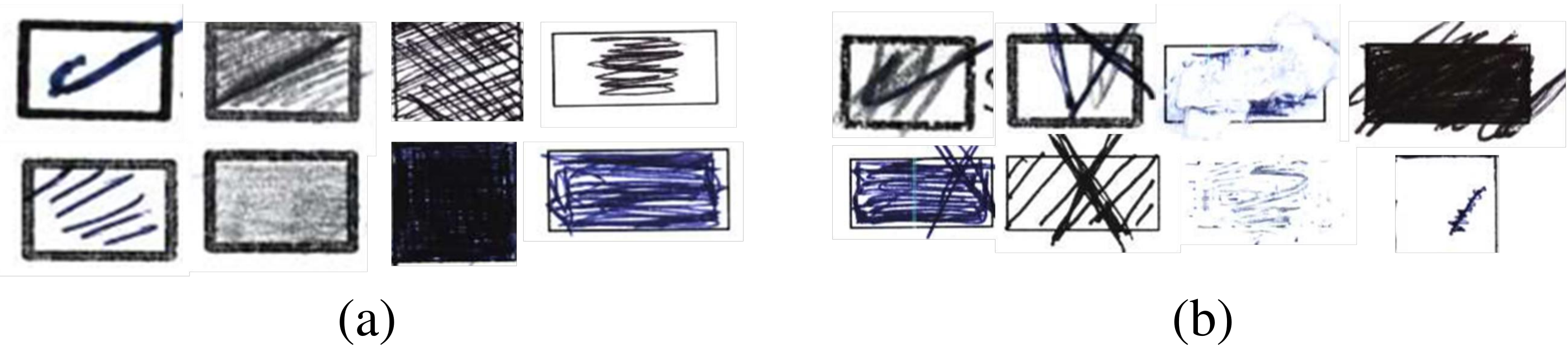}
\caption{Samples of (a) confirmed and (b) crossed out answers in the proposed dataset.}
\label{fig3}
\end{figure*}

\begin{figure*}
\centering
\includegraphics[width=0.7\linewidth]{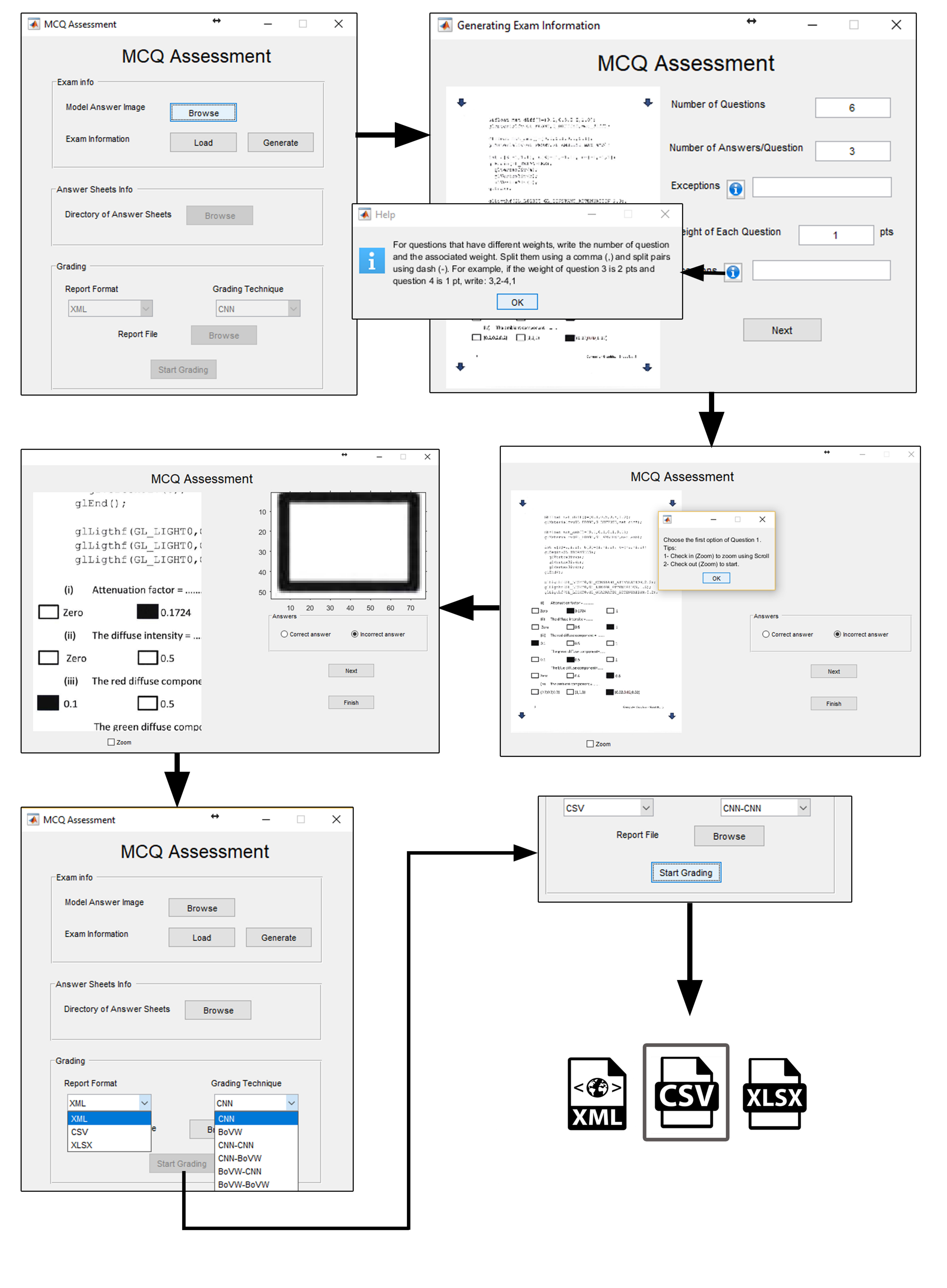}
\caption{Screenshots of the graphical user interface of the proposed system. First, the user uploads the model answer image and determines the metadata of it (e.g., number of questions, ROIs of answer boxes, and marks per question). Then, the user determines the directory of the images of the answer sheets. Eventually, the system automatically grades them and reports the results in one of three different formats: XML format (.XML), comma-separated values (.CSV), and Microsoft Excel workbook (.XLSX). Each record is presented as two columns: the image file name and the grade.}
\label{fig6}
\end{figure*}

The dataset comprises six real MCQ assessments. For each exam, a model answer sheet is supported. The MCQ exams have different answer sheet templates that are shown in Fig. \ref{fig1}. The students were informed that the assessment process will be done manually with a space of tolerance with the faint marks and crossed out marks. This provides us a set of different styles of confirmed and crossed out answers as shown in Fig. \ref{fig3}. Table \ref{table1} illustrates the details of all the exams. The documents had been scanned using a HP Scanjet Enterprise Flow N9120 Flatbed Scanner. The scanned documents had been saved in XML paper specification (XPS) file format, thereafter the XPS documents were converted to PNG images. The dataset is available in the following link: \href{https://goo.gl/q9AgHC}{https://goo.gl/q9AgHC}.

The metadata of each exam was created manually by 7 volunteers. The ROIs of answer boxes were determined in a semi-automated way by specifying the answer boxes on the attached model answer sheet. Then the scanned answer sheets were aligned to the reference image (i.e., the image of the model answer sheet) to avoid any misalignments during the scanning process. The volunteers determined the type of each extracted ROI (i.e., answer box) whether it is confirmed, crossed out (canceled), or empty (blank) answer. There is an additional information reported for each exam, such as the correct answer, the student ID location, number of pages, the page number of the current PNG image, and the grade of each answer sheet.

\subsection{Formating}
As mentioned earlier, the dataset comes in two formats: XPS documents (13 XPS files) and PNG images. Each PNG image is a scanned image of an individual page of each answer sheet. The filename contains the exam number, the number of the answer sheet, and the page number. For example, \texttt{exam0\_13\_2} refers to page number 2 of answer sheet number 13 of \textit{exam0}. The metadata describes each answer sheet by a set of variables for which the details are shown in Table \ref{table3}.

\begin{table*}
\centering
\caption{Description of the provided metadata of the proposed dataset.}
\label{table3}
\begin{tabular}{|l|p{12cm}|}
\hline
\textbf{Variable} & \textbf{Description} \\ \hline
\textit{imageName} & The filename of the associated image. The formatting composes of the exam ID, answer sheet number, and page number. \\  \hline
\textit{studentId} &  ID number associated with each student. \\ \hline
\textit{studentIdRect} & The ROI of the student ID in the answer sheet. \\ \hline
\textit{questionWeight} & The weight of each question, used for grading purposes. \\ \hline
\textit{totalNumberOfQuestions} & The total number of questions in the MCQ exam associated with the current image file.  \\ \hline
\textit{numberOfQuestionsPerPage} & The number of questions in the current page. \\ \hline
\textit{examNumberOfPages} &  The number of pages of the current MCQ exam.  \\ \hline
\textit{isThereAStudentId} &  Logical variable that refers whether the student ID is supported or not. \\  \hline
\textit{examId} & An integer that takes values from 0 to 5 indicating the examId. For example, when $examId=0$ means the associated image file is the answer sheet of \textit{exam0}.  \\ \hline
\textit{pageNumber} & The current page number. \\ \hline
\textit{answerType} & An integer that refers to the type of the corresponding ROI of the answer: 1 refers to a confirmed answer, 2 refers to a crossed out (cancelled) answer, and 3 refers to an empty (blank) answer box. \\ \hline
\textit{questionAnswer} & The correct answer of each question. \\ \hline
\textit{questionChoices} & The available choices for each question. \\ \hline
\textit{questionRect} &  The ROIs of the answer boxes visible in the current page. \\ \hline
\end{tabular}
\end{table*}

\section{Experimental Results}
\label{results}
In order to evaluate the accuracy of the image classification methods and the presented semi-automated assessment system, a set of experiments was carried out using the proposed dataset. We used Matlab 9.2 for the implementation of the presented methods 
The experiments were done on an Intel\textsuperscript{\textregistered} core\textsuperscript{TM} i-7 6700 @ 3.40GHz machine with 16 GB RAM and NVIDIA\textsuperscript{\textregistered} Quadro K620 graphics card. The experimental results show that the proposed system can handle crossed out answers, and different mark patterns and templates efficiently.
\subsection{Classification Results}
We have used 5-fold cross-validation to report the classification accuracy obtained by the examined image classification techniques. To increase the number of training crossed out samples, we have applied a set of transformations to each crossed out image. The transformations are: (1) translation by $p \in \{1,2,3,4\}$ pixels from left, right, bottom, and top, (2) rotation by $r \in\{-3^{\circ},-2^{\circ},..,2^{\circ},3^{\circ }\}$, and (3) flipping the image vertically and horizontally. This process increases the number of crossed out images in each fold from 40 to 680 images. Thus, we have 2,720 images for training and 680 for testing. In the BoVW, we have used 200 clusters to generate the pool of visual words. In both the NBC and CNN, we have resized all ROIs to ($227\times 227$ pixels) before training and testing stage. Table \ref{Table4} shows the average accuracy obtained by each of the three discussed methods.

The baseline model refers to training an SVM classifier using a simple feature vector representing the answer box. The feature vector consists of the mean intensity of the answer box for each color channel (i.e., the mean of the green color channel, the mean of the red color channel, and the mean of the blue color channel).

As shown, the best classification accuracy is achieved by the CNN-based solution. Where, the CNN-based classifier obtains 92.66\% accuracy using 3 classes. In the binary classification, the BoVW obtained the best accuracy (100\%) in confirmed and empty classes. CNN achieves the best accuracy for classifying between (confirmed and crossed out), and (crossed out and empty) answer boxes (91.41\% and 98.35\%, respectively).

As a matter of fact, the conflict comes from the similarity between confirmed and crossed out answers; as shown in the binary classification results, it is the lower accuracy obtained when the classifier is trained to distinguish between confirmed and crossed out answer boxes. In spite of the lowest results obtained by the NBC, it gets 91\% accuracy to recognize the confirmed answers in the traditional binary classification of confirmed and blank answers. For more results using different evaluation metrics, please see Table \ref{detailed_results}.

\subsection{Grading Accuracy}
To evaluate the grading accuracy of the proposed system, we have used the straight-forward and the two-stage classification methods. In the two-stage classification method, all permutations of the three classifiers, discussed earlier, have been examined. For each classification method, we have used the trained model that achieves the best score through the 5-fold cross validation. For each answer sheet, the ROIs are extracted and classified as described in Algorithms \ref{algo1} and \ref{algo2}. Two different evaluation criteria were used. The first one is question-based, where the reported accuracy is given by 
\begin{equation}
\text{accuracy}=\frac{\text{questions corrected properly}}{\text{total number of questions}}.
\end{equation}
The second criterion is answer sheet-based approach. The accuracy is calculated by
\begin{equation}
\label{eq1}
\text{accuracy}=\frac{\text{answer sheets corrected properly}}{\text{total number of answer sheets}}.
\end{equation}
As shown, the second criterion is more strict. For example, if only one question is incorrectly graded, the entire answer sheet is not counted in the numerator of Equation \ref{eq1}. Eventually, we reported the average accuracy of the question-based and the answer sheet-based evaluation criteria.

For comparisons, we have applied the recognition methods presented by Deng et al. \cite{denglow} and Sanguansat \cite{sanguansatrobust} to the proposed dataset.  
According to Table \ref{Table5}, the end-to-end CNN classifier, namely the 3 classes classifier, attains the best accuracy in both evaluation criteria by achieving 99.78\% and 97.69\% for the first and second evaluation criteria, respectively. However, the two-stage classification strategy improves the accuracy obtained by the BoVW, where the BoVW-based straight-forward classifier obtains 92.3\% and 68.6\% for question-based and answer sheet-based assessment criteria, respectively. That is improved by +6.7\% and +24.87\%, respectively, using the two-stage classification strategy.

It is clear that the NBC attains the worst results either in the straight-forward or two-stage classification strategies. That, however, is improved by involving either the BoVW or CNN classifiers in the two-stage classification process. It is obvious that the simple thresholding process, performed in most of the existing systems, is insufficient to deal with our dataset because of the variations in the mark patterns, and the crossed out answer boxes.

 \begin{table}
 \centering
 \caption{Average classification accuracy obtained using 5-fold cross-validation. The target classes are: (a) confirmed, crossed out, and empty, (b) confirmed and empty, (c) confirmed and crossed out, and (d) crossed out and empty.}
 \label{Table4}
  \scalebox{0.9}{
 \begin{tabular}{|c|c|c|c|c|}
 \hline
 \multirow{2}{*}{\textbf{Classifiers}} & \multicolumn{4}{c|}{\textbf{Classes}} \\ \cline{2-5} 
  & \textbf{(a)} & \textbf{(b)} & \textbf{(c)} & \textbf{(d)} \\ \hline
 \textbf{Baseline} & 0.604 & 0.825 & 0.503 & 0.5\\ \hline
 \textbf{NBC} & 0.293 & 0.91 & 0.478 & 0.797 \\ \hline
 \textbf{BoVW} & 0.909 & \textbf{1} & 0.891 &\textbf{ 0.97} \\ \hline
 \textbf{CNN} &\textbf{ 0.927} &\textbf{ 0.999} & 0.9141 &\textbf{ 0.984 }\\ \hline
 \end{tabular}
 }
 \end{table}

 \begin{table}[]
 \centering
 \caption{Precision, recall, and F score for the bag of visual words (BoVW) and the CNN models.}
 \label{detailed_results}
 \scalebox{0.72}{
 \begin{tabular}{l|l|l|l|l|l|l|}
 \cline{2-7}
  & \multicolumn{3}{c|}{confirmed} & \multicolumn{3}{c|}{crossed out} \\ \hline
 \multicolumn{1}{|l|}{Method} & Precision & Recall & F score & Precision & Recall & F score \\ \hline
 \multicolumn{1}{|l|}{BoVW} & 0.91 & 0.914 & 0.912 & 0.911 & 0.844 & 0.876 \\ \hline
 \multicolumn{1}{|l|}{CNN} & 0.829 & 0.999 & 0.906 & 0.998 & 0.79 & 0.88 \\ \hline
 \end{tabular}}
 \end{table}

  \begin{table}[t]
  \centering
 \caption{Grading accuracy of the proposed system. SF refers to straight-forward classification strategy. The term 2S refers to the two-stage classifier. The order of each classification method, in the case of (2S), indicates the order of the classifier methods used in the two-stage classifier. For example BoVW-NBC (2S), means that the first classifier (confirmed and empty classifier) is the BoVW and the second classifier (confirmed and crossed out) is the NBC.}

 \label{Table5}
 \scalebox{0.85}{
 \begin{tabular}{|c|c|c|}
 \hline
 \multirow{2}{*}{\begin{tabular}[c]{@{}c@{}}\\   Method\end{tabular}} & \multicolumn{2}{c|}{Accuracy} \\ \cline{2-3} 
  & Question-based & Answer sheet-based \\ \hline
    H. Deng et al. \cite{denglow} &0.43
      & 0.05
       \\ \hline
   P. Sanguansat \cite{sanguansatrobust} 
    & 0.32 & 0.08
     \\ \hline
 BoVW (SF) & 0.9230 & 0.6860 \\ \hline
 CNN (SF) & \textbf{0.9978} & \textbf{0.9769} \\ \hline
 NBC (SF) & 0.7801 & 0.0803 \\ \hline
NBC-NBC (2S) & 0.7384 & 0.0422 \\ \hline
NBC-BoVW (2S) & 0.9190 & 0.5778 \\ \hline
NBC-CNN (2S) & 0.8965 & 0.4612 \\ \hline
 BoVW-BoVW (2S) & \textbf{0.9926} & \textbf{0.9347} \\ \hline
 BoVW-NBC (2S) & \textbf{0.9889} & 0.9007 \\ \hline
 BoVW-CNN (2S) & \textbf{0.9502} & \textbf{0.9551} \\ \hline
 CNN-CNN (2S) & \textbf{0.9978} & \textbf{0.9769} \\ \hline
 CNN-NBC (2S) & \textbf{0.9896} & 0.8980 \\ \hline
 CNN-BoVW (2S) & \textbf{0.9939} & \textbf{0.9402} \\ \hline
 
 \end{tabular}
}
 \end{table}

 \begin{table*}
 \centering
 \caption{The classification accuracies of the cross-dataset evaluation. We used a new dataset to test the generalization of our trained models. The new dataset contains different answer boxes' shapes and has been scanned using a different scanner. The order of each classification method, in the case of (2S), indicates the order of the classifier methods used in the two-stage classifier. For example BoVW-CNN (2S), means that the first classifier (confirmed and empty classifier) is the BoVW and the second classifier (confirmed and crossed out) is the CNN.}
 \label{set2_scanner}
 \scalebox{0.9}{
 \begin{tabular}{ll|c|c|c|c|c|c|c|}
 \cline{3-9}
  &  & \multicolumn{7}{c|}{Methods} \\ \cline{3-9} 
  &  & BoVW (SF) & CNN (SF) & BoVW (2S) & CNN (2S) & CNN-BoVW (2S) & BoVW-CNN (2S) & Baseline model \\ \hline
 \multicolumn{2}{|l|}{Accuracy} & 0.743 & 0.631 & 0.711 & 0.699 & 0.573 &\textbf{ 0.933} & 0.488 \\ \hline
 \end{tabular}
 }
 \end{table*}

 \begin{table*}
 \centering
 \caption{Precision, recall, and F score of the cross-dataset evaluation. The order of each classification method, in the case of (2S), indicates the order of the classifier methods used in the two-stage classifier. For example BoVW-CNN (2S), means that the first classifier (confirmed and empty classifier) is the BoVW and the second classifier (confirmed and crossed out) is the CNN.}
 \label{set2_more}
 \scalebox{0.8}{
 \begin{tabular}{ll|c|c|c|c|c|c|c|}
 \cline{3-9}
  &  & \multicolumn{7}{c|}{Methods} \\ \cline{3-9} 
  &  & BoVW (SF) & CNN (SF) & BoVW (2S) & CNN (2S) & CNN-BoVW (2S) & BoVW-CNN (2S) & Baseline model \\ \hline
 \multicolumn{1}{|l|}{\multirow{3}{*}{Confirmed}} & Precision & 1 & 0.5 & 1 & 0.593 & 0.340 & 0.863 & 0.5 \\ \cline{2-9} 
 \multicolumn{1}{|l|}{} & Recall & 0.36 & 1 & 0.28 & 1 & 0.28 & 0.99 & 0.013 \\ \cline{2-9} 
 \multicolumn{1}{|l|}{} & F score & 0.529 & 0.667 & 0.438 & 0.744 & 0.307 & 0.922 & 0.03 \\ \hline
 \multicolumn{1}{|l|}{\multirow{3}{*}{Crossed out}} & Precision & 0.446 & 0.658 & 0.41 & 0.849 & 0.392 & 1 & 0.674 \\ \cline{2-9} 
 \multicolumn{1}{|l|}{} & Recall & 1 & 0.649 & 1 & 0.682 & 1 & 0.682 & 0.419 \\ \cline{2-9} 
 \multicolumn{1}{|l|}{} & F score & 0.617 & 0.653 & 0.582 & 0.757 & 0.563 & 0.811 & 0.517 \\ \hline
 \end{tabular}
 }
 \end{table*}

 \begin{table*}[!t]
 \centering
 \caption{The classification accuracies of the cross-dataset evaluation captured using mobile phone cameras. The new dataset contains different answer boxes' shapes and has been captured using two different mobile phone cameras. The order of each classification method, in the case of (2S), indicates the order of the classifier methods used in the two-stage classifier. For example BoVW-CNN (2S), means that the first classifier (confirmed and empty classifier) is the BoVW and the second classifier (confirmed and crossed out) is the CNN.}
 \label{set2_mobile}
 \scalebox{0.9}{
 \begin{tabular}{ll|c|c|c|c|c|c|c|}
 \cline{3-9}
  &  & \multicolumn{7}{c|}{Methods} \\ \cline{3-9} 
  &  & BoVW (SF) & CNN (SF) & BoVW (2S) & CNN (2S) & CNN-BoVW (2S) & BoVW-CNN (2S) & Baseline model \\ \hline
 \multicolumn{2}{|l|}{Accuracy} & 0.717 & 0.602 & 0.708 & 0.655 & 0.551 &\textbf{ 0.909} & 0.417 \\ \hline
 \end{tabular}
 }
 \end{table*}

 \begin{table*}[!t]
 \centering
 \caption{Average CPU execution time of the image classification methods. The term SF refers to straight-forward classification strategy. The term 2S refers to the two-stage classifier. The order of each classification method, in the case of (2S), indicates the order of the classifier methods used in the two-stage classifier.}
 \label{timeAnalysis}
  \scalebox{0.86}{
 \begin{tabular}{|c|c|c|}
 \hline
 \multirow{2}{*}{Method} & \multicolumn{2}{c|}{Average execution time on CPU (ms)} \\ \cline{2-3} 
  & Answer box & Question (3 options) \\ \hline
 BoVW (SF) & 21.7 & 65.1 \\ \hline
 CNN (SF) & 526.7 & 1580.1 \\ \hline
 NBC (SF) & 3.6 &  10.8 \\ \hline
 NBC-Naive (2S) & 4.5 & 13.5  \\ \hline
 NBC-BoVW (2S) & 10.4 & 31.2 \\ \hline
 NBC-CNN (2S) & 161.9 & 485.7 \\ \hline
 BoVW-BoVW (2S) & 29.8 &  89.4 \\ \hline
 BoVW-NBC (2S) & 23.9 &  71.7 \\ \hline
 BoVW-CNN (2S) & 181.3 & 543.9 \\ \hline
 CNN-CNN (2S) & 686.5 & 2059.5  \\ \hline
 CNN-NBC (2S) & 529.2 & 1587.6 \\ \hline
 CNN-BoVW (2S) & 535.2 & 1605.6 \\ \hline
 \end{tabular}
 }
 \end{table*}

 \subsection{Generalization} \label{generalizationSec}
Our trained models, the CNN and the BoVW, give promising classification accuracies between the three different types of answer boxes, and consequently, the achieved grading accuracies outperform the results obtained by the simple threshold-based methods. In spite of the good results obtained, we are interested in testing the generalization of our models -- how well it performs on data that has been collected using different devices or has different shapes of answer boxes.

Essentially, the BoVW model is based on affine invariant features which are robust against the affine transformations. These feature vectors represent different patches from the answer box (i.e., the salient image blocks). Thus, the feature vector is supposed to be robust against changing the answer boxes' shapes.

For our trained CNN model, we have applied a dropout layer with 50\% drop-out rate after each fully connected layer, as described in Table \ref{AlexNetTable}. That means we use only 50\% of the fully connected layers' capacities aiming to prevent overfitting.

It is clear that classifying confirmed answers and the empty answers (i.e., filled vs. unfilled) is an easy task. The difficulty comes when we try to distinguish between crossed out and the confirmed answers, because there is similarity between their patterns. For that reason, there is a potential overfitting associated with the complex models, (e.g., the CNN model). 

\begin{figure*}
 \centering
 \includegraphics[width=\linewidth]{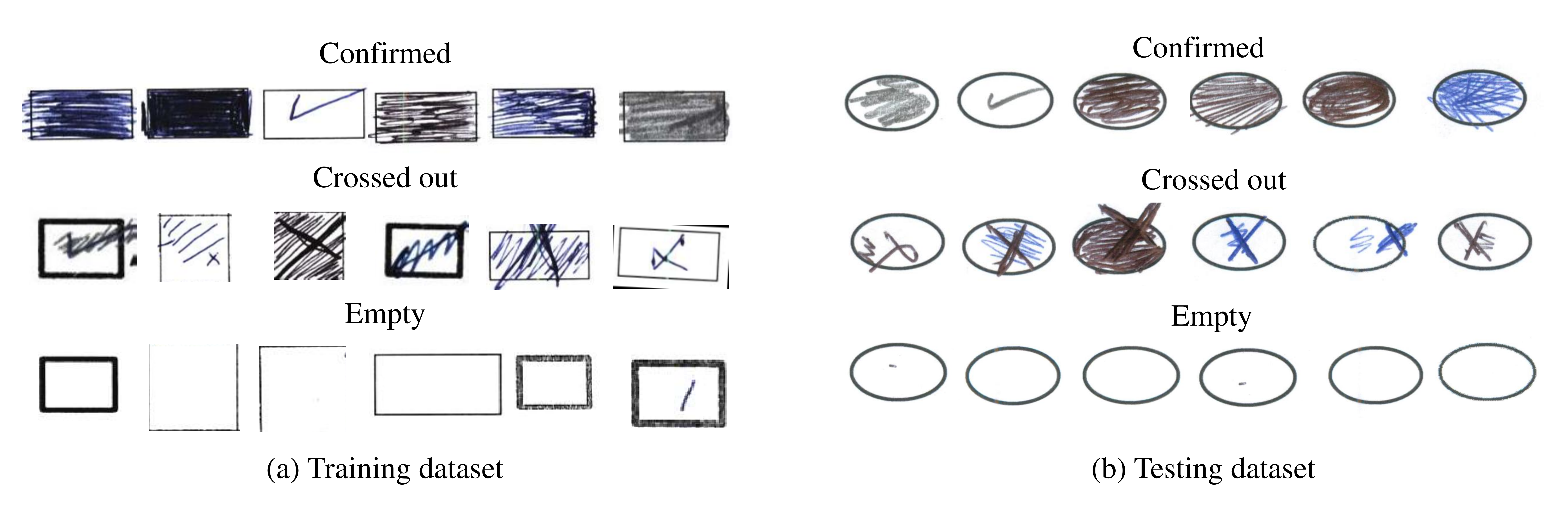}
 \caption{Samples of the data used for the cross-dataset evaluation. (a) Samples of the answer boxes used to train the classifiers from our proposed dataset in Sec. \ref{dataset}. (b) Samples of the answer boxes used for testing.}
 \label{twodatasets}
 \end{figure*}

In order to test the generalization of the trained models, we have collected a new set of answer boxes that has been obtained using a different scanner. Our new set has been scanned using Fuji ScanSnap IX500 -- a different scanner than what we used to scan our dataset (see Sec. \ref{dataset}). Specifically, by using two different acquisition devices, the new set is sampled from a new distribution than the distribution of the training set \cite{storkey2009training}. Furthermore, the answer box shapes are oval boxes instead of the rectangular answer boxes that were used in the original dataset. The total numbers of confirmed, crossed out, and empty answer boxes are 300, 148, and 298, respectively. We have used this new set to test the trained models on our dataset (presented in Sec. \ref{dataset}). Fig. \ref{twodatasets} shows samples of the answer boxes that have been used to train the models and another samples of the answer boxes from the new set.

The results of the cross-dataset evaluation are shown in Table \ref{set2_scanner}. As shown, the BoVW and CNN models outperform the baseline model. However, the achieved accuracies dropped down significantly for the end-to-end model indicating that the trained model is not generalizing well for other datasets with other answer boxs' shapes. The best results are obtained using the two-stage classifier (93.3\% accuracy) where the BoVW is used to classify the empty and confirmed answer boxes, and the CNN is used to classify between the confirmed and the crossed out answer boxes.

The comparison between the precision, recall and the F score achieved by the two-stage classifier (i.e., BoVW-CNN) against the values obtained by the end-to-end CNN model and the 3-classes BoVW model is shown in Table \ref{set2_more}. The results obtained using the cross-dataset evaluation -- namely, the testing dataset is different from what we have used to train the models in terms of scanning device and shapes of the answer boxes, show that by dividing the problem into two stages, we could benefit from the best of each model. The CNN model is considered the best model to distinguish between the crossed out and confirmed answers; while the BoVW is more effective to classify confirmed and empty answers, because it is less sensitive to overfitting.

\subsection{Testing on Mobile Phone Images}
For the sake of completeness, we have examined the trained models on the same new set (introduced in Sec. \ref{generalizationSec}), but the answer sheets were captured by two different mobile phone cameras. 

We used the same mobile phone models used in our experiments in Sec. \ref{existing} under similar lighting conditions. The obtained accuracy of each model is shown in Table \ref{set2_mobile}. As shown, the two-stage classifier BoVW-CNN achieves the best results (90.9\%). 

 \subsection{Time Analysis}
One important issue of using CNN solutions is the computation complexity. Undoubtedly, the GPU-accelerated CNN architecture reduces the time required to learn the big number of CNN's parameters. 

In order to get a fair comparison, we used the CPU-based implementation of the CNN architecture \cite{alexnet} to compare the time required by all image classification methods discussed in this work. Note that the GPU-based CNN implementation runs $\texttildelow$\textit{ 20 times} faster than the CPU version presented below.

From the performance aspects, the CNN is considered the slowest step in the two-stage classifiers. The straight-forward classification using a CNN is considered the slowest straight-forward classifier. The NBC, in both straight-forward and two-stage classifiers, requires the shortest time. Table \ref{timeAnalysis} shows more details of the time analysis of the image classification methods discussed in this paper. As shown, there is a direct correlation between the complexity level of the image classification methods (see Fig. \ref{summary}) and the inference time.

\section{Conclusions and Future Work} \label{conclusion}
In this work, we have proposed a semi-automated OMR system that aims to reduce the restrictions in MCQ tests. The recognition of answer boxes is considered a crucial step in any OMR system, nonetheless, the existing systems use simple image processing techniques that lead to a set of restrictions in the MCQ tests. In this work, however, we have shown that by using a more sophisticated ML-based approach, high accurate OMR systems can be developed without restrictions on the marking of the answer boxes, e.g., crossing out wrong answers. We have examined three image classification methods: (1) NBC, (2) BoVW, and (3) CNN, in two different classification strategies to get the best potential result. In order to train each model, we have proposed a new dataset including 735 answer sheets collected from six different MCQ assessments. We have shown that by splitting the problem into two binary classification problems, the accuracy of the BoVW and NBC is improved. We have found that the end-to-end CNN solution is considered the best by correcting 99.39\% of the questions properly. In terms of generalization, however, we have found that the two-stage strategy for classification can generalize better than the end-to-end CNN solution achieving 93.3\% and 90.9\% accuracy rates in the cross-dataset evaluation using another scanner device and mobile phone cameras, respectively. Also, the cross-dataset evaluation includes different answer box shapes.

Although ML-based classifiers attain good accuracy in dealing with crossed out answer boxes, an effective procedure for error detection and correction is still needed. Future work would be to provide a mechanism to locate and correct
the errors quickly and easily to ensure 100\% accuracy of the final reported result. Other directions include examining recurrent neural network (RNN) to provide an end-to-end process to recognize the selected answer among available options for each question.


\begin{thebibliography}{10}

\bibitem{gronlundassessment}
Norman~E Gronlund.
\newblock {\em Assessment of student achievement}.
\newblock ERIC, 1998.

\bibitem{mccoubrieimproving}
Paul McCoubrie.
\newblock Improving the fairness of multiple-choice questions: a literature
  review.
\newblock {\em Medical teacher}, 26(8):709--712, 2004.

\bibitem{aldabesemantic}
Itziar Aldabe and Montse Maritxalar.
\newblock Semantic similarity measures for the generation of science tests in
  basque.
\newblock {\em IEEE transactions on Learning Technologies}, 7(4):375--387,
  2014.

\bibitem{liuautomatic}
Ming Liu, Vasile Rus, and Li~Liu.
\newblock Automatic chinese multiple choice question generation using mixed
  similarity strategy.
\newblock {\em IEEE Transactions on Learning Technologies}, 11(2):193--202, 2017.

\bibitem{spadaccinimultiple}
Andrea Spadaccini and Vanni Rizzo.
\newblock A multiple-choice test recognition system based on the gamera
  framework.
\newblock {\em arXiv preprint arXiv:1105.3834}, 2011.

\bibitem{chaiautomated}
Douglas Chai.
\newblock Automated marking of printed multiple choice answer sheets.
\newblock In {\em IEEE International Conference on Teaching, Assessment, and Learning for Engineering (TALE)}, pages 145--149, 2016.

\bibitem{fisteusgrading}
Jesus~Arias Fisteus, Abelardo Pardo, and Norberto~Fern{\'a}ndez Garc{\'\i}a.
\newblock Grading multiple choice exams with low-cost and portable
  computer-vision techniques.
\newblock {\em Journal of Science Education and Technology}, 22(4):560--571,
  2013.

\bibitem{ahmedocr}
Ali~H Ahmed, Mahmoud Afifi, Mostafa Korashy, Ebram~K William, Mahmoud~Abd El-sattar, and Zenab Hafez.
\newblock OCR system for poor quality images using chain--code representation.
\newblock In {\em The 1st International Conference on Advanced Intelligent
  System and Informatics}, pages 151--161, 2016.

\bibitem{wangend}
Tao Wang, David~J Wu, Adam Coates, and Andrew~Y Ng.
\newblock End-to-end text recognition with convolutional neural networks.
\newblock In {\em 21st International
  Conference on Pattern Recognition (ICPR)}, pages 3304--3308, 2012.

\bibitem{jaderbergreading}
Max Jaderberg, Karen Simonyan, Andrea Vedaldi, and Andrew Zisserman.
\newblock Reading text in the wild with convolutional neural networks.
\newblock {\em International Journal of Computer Vision}, 116(1):1--20, 2016.

\bibitem{chinnasarnimage}
Krisana Chinnasarn and Yuttapong Rangsanseri.
\newblock An image-processing oriented optical mark reader.
\newblock {\em J. Society of Photo-Optical Instrumentation Engineers}, 3808:702--709, 1999.

\bibitem{nguyenefficient}
Tien~Dzung Nguyen, Quyet~Hoang Manh, Phuong~Bui Minh, Long~Nguyen Thanh, and
  Thang~Manh Hoang.
\newblock Efficient and reliable camera based multiple-choice test grading
  system.
\newblock In {\em International Conference on Advanced Technologies for Communications (ATC)}, pages 268--271, 2011.

\bibitem{hussmannhigh}
Stephan Hussmann and Peter~Weiping Deng.
\newblock A high-speed optical mark reader hardware implementation at low cost
  using programmable logic.
\newblock {\em Real-Time Imaging}, 11(1):19--30, 2005.

\bibitem{denglow}
Hui Deng, Feng Wang, and Bo~Liang.
\newblock A low-cost {O}{M}{R} solution for educational applications.
\newblock In {\em International Symposium on Parallel and Distributed Processing with Applications}, pages 967--970, 2008.

\bibitem{levimethod}
Joseph~A Levi, Yosef~A Solewicz, Yael Dvir, and Yisca Steinberg.
\newblock Method of verifying declared identity in optical answer sheets.
\newblock {\em Soft Computing}, 15(3):461--468, 2011.

\bibitem{chouvatutflexible}
Varin Chouvatut and Supachaya Prathan.
\newblock The flexible and adaptive x-mark detection for the simple answer
  sheets.
\newblock In {\em International Computer Science and Engineering Conference (ICSEC)}, pages 433--439, 2014.

\bibitem{maaten2008visualizing}
Laurens van~der Maaten and Geoffrey Hinton.
\newblock Visualizing data using t-sne.
\newblock {\em Journal of machine learning research}, 9:2579--2605, 2008.

\bibitem{otsuthreshold}
Nobuyuki Otsu.
\newblock A threshold selection method from gray-level histograms.
\newblock {\em IEEE transactions on systems, man, and cybernetics},
  9(1):62--66, 1979.

\bibitem{new1}
Daniel Lopresti, George Nagy, and Elisa~Barney Smith.
\newblock A document analysis system for supporting electronic voting research.
\newblock In {\em The Eighth IAPR International Workshop on Document Analysis Systems}, pages 167--174, 2008.

\bibitem{new2}
Elisa~H Barney-Smith, George Nagy, and Daniel Lopresti.
\newblock Mark detection from scanned ballots.
\newblock In {\em Document Recognition and Retrieval XVI}, volume 7247, page
  72470P-1--72470P-11. International Society for Optics and Photonics, 2009.

\bibitem{new3}
Elisa H~Barney Smith, Daniel Lopresti, George Nagy, and Ziyan Wu.
\newblock Towards improved paper-based election technology.
\newblock In {\em International
  Conference on Document Analysis and Recognition (ICDAR)}, pages 1255--1259, 2011.

\bibitem{new4}
George Nagy and Daniel Lopresti.
\newblock The role of document image analysis in trustworthy elections.
\newblock In {\em Advances in Digital Document Processing and Retrieval}, pages
  51--81, 2014.

\bibitem{new5}
Elisa H~Barney Smith, Daniel Lopresti, George Nagy, and Ziyan Wu.
\newblock Towards improved paper-based election technology.
\newblock In {\em International
  Conference on Document Analysis and Recognition (ICDAR)}, pages 1255--1259, 2011.

\bibitem{smithoptical}
Andrew~M Smith.
\newblock Optical mark reading-making it easy for users.
\newblock In {\em Proceedings of the 9th annual ACM SIGUCCS conference on User
  services}, pages 257--263, 1981.

\bibitem{loprestidocument}
Daniel Lopresti, George Nagy, and Elisa~Barney Smith.
\newblock A document analysis system for supporting electronic voting research.
\newblock In {\em The Eighth International Workshop on Document Analysis Systems}, pages 167--174, 2008.

\bibitem{sanguansatrobust}
Parinya Sanguansat.
\newblock Robust and low-cost optical mark recognition for automated data
  entry.
\newblock In {\em International Conference on Electrical Engineering/Electronics, Computer, Telecommunications and Information Technology (ECTI-CON)}, pages 1--5, 2015.

\bibitem{chinaapplication}
Rodrigo~Teiske China, Francisco de~Assis~Zampirolli, Rog{\'e}rio~Perino
  de~Oliveira~Neves, and Jos{\'e}~Artur Quilici-Gonzalez.
\newblock An application for automatic multiple-choice test grading on android.
\newblock {\em Revista Brasileira de Inicia{\c{c}}{\~a}o Cient{\'\i}fica}, 3(2):4–-25, 2016.

\bibitem{SURF}
Herbert Bay, Tinne Tuytelaars, and Luc Van~Gool.
\newblock Surf: Speeded up robust features. \newblock {\em ECCV}, pages 404--417, 2006.

\bibitem{matching}
Marius Muja and David~G Lowe.
\newblock Fast matching of binary features.
\newblock In {\em Ninth Conference on Computer and Robot Vision (CRV)}, pages 404--410, 2012.

\bibitem{MSAC}
Philip~HS Torr and David~W Murray.
\newblock The development and comparison of robust methods for estimating the
  fundamental matrix.
\newblock {\em International journal of computer vision}, 24(3):271--300, 1997.

\bibitem{naive}
David~D Lewis.
\newblock Naive (bayes) at forty: The independence assumption in information
  retrieval.
\newblock In {\em European conference on machine learning}, pages 4--15, 1998.

\bibitem{bovw}
Gabriella Csurka, Christopher Dance, Lixin Fan, Jutta Willamowski, and
  C{\'e}dric Bray.
\newblock Visual categorization with bags of keypoints.
\newblock In {\em Workshop on statistical learning in computer vision, ECCV},
pages 1--2, 2004.

\bibitem{alexnet}
Alex Krizhevsky, Ilya Sutskever, and Geoffrey~E Hinton.
\newblock Imagenet classification with deep convolutional neural networks.
\newblock In {\em Advances in neural information processing systems}, pages
  1097--1105, 2012.

\bibitem{HOG}
Navneet Dalal and Bill Triggs.
\newblock Histograms of oriented gradients for human detection.
\newblock In {\em CVPR}, pages 886--893, 2005.

\bibitem{svm}
Vladimir~Naumovich Vapnik and Vlamimir Vapnik.
\newblock {\em Statistical learning theory},
\newblock Wiley New York, 1998.

\bibitem{harr}
Krystian Mikolajczyk and Cordelia Schmid.
\newblock An affine invariant interest point detector.
\newblock {\em ECCV}, pages 128--142, 2002.

\bibitem{duda}
Richard~O Duda, Peter~E Hart, and David~G Stork.
\newblock {\em Pattern classification}.
\newblock John Wiley \& Sons, 2012.

\bibitem{deep}
Yann LeCun, Yoshua Bengio, and Geoffrey Hinton.
\newblock Deep learning.
\newblock {\em Nature}, 521(7553):436--444, 2015.

\bibitem{vgg}
Karen Simonyan and Andrew Zisserman.
\newblock Very deep convolutional networks for large-scale image recognition.
\newblock {\em arXiv preprint arXiv:1409.1556}, 2014.

\bibitem{resnet}
Kaiming He, Xiangyu Zhang, Shaoqing Ren, and Jian Sun.
\newblock Deep residual learning for image recognition.
\newblock In {\em CVPR}, pages 770--778, 2016.

\bibitem{canzianianalysis}
Alfredo Canziani, Adam Paszke, and Eugenio Culurciello.
\newblock An analysis of deep neural network models for practical applications.
\newblock {\em arXiv preprint arXiv:1605.07678}, 2016.

\bibitem{murphymachine}
Kevin~P Murphy.
\newblock {\em Machine learning: a probabilistic perspective}.
\newblock MIT press, 2012.

\bibitem{storkey2009training}
Amos Storkey.
\newblock When training and test sets are different: characterizing learning
  transfer.
\newblock {\em Dataset shift in machine learning}, pages 3--28, 2009.

\end{thebibliography}
\end{document}